\pdfoutput=1
\documentclass[11pt]{article}
\usepackage[review]{ACL2023}

\usepackage{times}
\usepackage{latexsym}

\usepackage[T1]{fontenc}
\usepackage{microtype}

\usepackage{inconsolata}
\usepackage{graphicx}
\usepackage{booktabs}
\title{MedChatZH: a Better Medical Adviser Learns from Better Instructions}

\author{Yang Tan, Mingchen Li, Zijie Huang, Huiqun Yu and Guisheng Fan \\
  Department of Computer Science and Technology, \\
  East China University of Science and Technology, China \\
  \{tyang,lmc,hzj\}@mail.ecust.edu.cn \{yhq,gsfan\}@ecust.edu.cn}

\begin{document}
\maketitle
\begin{abstract}
Generative large language models (LLMs) have shown great success in various applications, including question-answering (QA) and dialogue systems. However, in specialized domains like traditional Chinese medical QA, these models may perform unsatisfactorily without fine-tuning on domain-specific datasets. To address this, we introduce MedChatZH, a dialogue model designed specifically for traditional Chinese medical QA. Our model is pre-trained on Chinese traditional medical books and fine-tuned with a carefully curated medical instruction dataset. It outperforms several solid baselines on a real-world medical dialogue dataset. We release our model, code, and dataset on https://github.com/tyang816/MedChatZH to facilitate further research in the domain of traditional Chinese medicine and LLMs.
\end{abstract}

\section{Introduction}

The ChatGPT series has achieved remarkable success in both academic and industrial circles, serving as a catalyst for numerous subsequent studies. Through a combination of instruction tuning and human feedback, these models have consistently demonstrated state-of-the-art performance across a wide range of Natural Language Processing (NLP) tasks. However, it is worth noting that these models are not openly available and do not divulge many specifics about their training process.

In recent years, several alternative foundational models have emerged in response to this limitation. For instance, LLaMa \cite{touvron2023llama}, BLOOM \cite{scao2022bloom}, and GLM \cite{du2021glm} are notable examples. These models have been trained on extensive collections of general raw texts derived from real-world sources, thereby introducing a new paradigm for comprehending fundamental knowledge within human society. By leveraging such diverse and expansive training data, these models offer unique insights and capabilities in understanding and processing natural language.

Given the constraints imposed by the limited availability of high-quality corpora, most Large Language Models (LLMs) are primarily tailored to cater to English-speaking users. Unfortunately, their performance significantly deteriorates when deployed in scenarios involving other languages. Furthermore, the performance of general-purpose large language models cannot be universally remarkable across various specialized domains \cite{zhang2023huatuogpt}. An illustrative example of this phenomenon lies in the commercialization of ChatGPT, which imposes certain restrictions on the provision of answers within the medical field. Consequently, a considerable disparity arises, wherein medical resources are scarce despite the limited scope of their application. This disconnect presents a challenge in terms of harnessing the full potential of these resources in the medical domain.

Our main contributions can be summarized as follows:
\begin{itemize}
    \item We enhanced the Chinese-specific language model by training it on an extensive collection of traditional Chinese medicine (TCM) books. As a result, the model is capable of providing answers that combine knowledge from both traditional Chinese and Western medicine.
    \item We curated a new dataset of medical dialogue instructions through a sophisticated pipeline that meticulously removed any irrelevant or sensitive data, such as private information and colloquial responses. 
    \item We demonstrated state-of-the-art performance on a real-world medical QA benchmark, outperforming other baseline models across several evaluation metrics. Furthermore, we have made our dataset and model open-source for the benefit of the research community.
\end{itemize}

\section{Related Work}

\subsection{Training General Language Models}

Training General language models consume trillion tokens and costly computation resources to learn the structure, syntax, and semantics of the human language through unsupervised methods. This stage allows the model to learn general language patterns and representations.

The Transformer \cite{vaswani2017attention} revolutionized natural language processing with its introduction of attention mechanisms, inspiring subsequent encoder-only architectures like BERT \cite{devlin2018bert} that leverage masked language modeling, as well as causal models such as the GPT \cite{radford2018gpt, radford2019gpt2, brown2020gpt3} series that utilize next token prediction strategy. However, since OpenAI releases ChatGPT and GPT-4, the casual language models have shown more potential power in modeling the real world, but their models' weights and training details are not open to the public. 

As alternatives, both LLaMa \cite{touvron2023llama} and BLOOM \cite{scao2022bloom} have released models' weights with more than 10 billion parameters for research purposes, but they focus on English applications and trained on massive English corpus. As alternatives, both LLaMa \cite{touvron2023llama} and BLOOM \cite{scao2022bloom} have made the weights of their models, each containing over 10 billion parameters, accessible for research purposes. However, their focus has primarily been on English applications, with training conducted on extensive English corpora. Recognizing the need to bridge the language gap in Chinese applications, ChatGLM \cite{du2021glm, zeng2022glm130b} employs an auto-regressive GLM with multiple training objectives and a bilingual corpus, achieving superior performance in Chinese-specific tasks. To address Chinese language requirements, TigerBot \footnote{https://github.com/TigerResearch/TigerBot} and BaiCahuan \footnote{https://github.com/baichuan-inc/baichuan-7B} have been developed based on the BLOOM and LLaMa architectures, respectively. These models are commercially available and cater to Chinese language processing needs.

\subsection{Medical Language Models}

While general-purpose Language Models (LMs) have demonstrated remarkable capabilities in various scenarios, it is often necessary to fine-tune them on specific, smaller datasets that are tailored to the target task or domain. This fine-tuning process helps the models to better understand and adapt to the specific requirements of downstream tasks.

In comparison to general-purpose models, specialized models for specific verticals are relatively scarce. For instance, BenTso \cite{wang2023huatuo} constructed a Chinese medical instruction dataset by leveraging the Medical Knowledge Graph and GPT3.5 API. Building upon this dataset, we performed fine-tuning on the instructions of LLaMA to enhance its query and answer effectiveness specifically in the medical field. The resulting model, HuatuoGPT \cite{zhang2023huatuogpt}, is a large language model trained on an extensive Chinese medical corpus, with the goal of constructing a more proficient 'ChatGPT' for medical consultation scenarios.

Additionally, Google's Med-PaLM \cite{singhal2022med-palm} harnesses the power of Google's large language models. These models have been aligned with the medical domain and evaluated using medical exams, medical research, and consumer queries in the English language. This alignment and evaluation process ensures that the model is well-suited for handling medical-related tasks and inquiries.

By developing and fine-tuning these specialized models, we aim to provide more accurate and reliable language processing solutions in domains such as healthcare and medicine. These models bridge the gap between general-purpose LMs and specific vertical applications, enabling more effective and targeted language understanding and generation in specialized fields.

\begin{table*}[]
\centering
\caption{Resulsts on webMedQA benchmark.  }
\label{tab:my-table}
\resizebox{\textwidth}{!}{%
\begin{tabular}{@{}llllllllll@{}}
\toprule
Model           & Parameter   & BLEU-1 & BLEU-2 & BLEU-3 & BLEU-4 & GLEU & ROUGE-1 & ROUGE-2 & ROUGE-L \\ \midrule
GPT-3.5-turbo * & /           & 18.06  & 6.74   & 2.73   & 1.09   & 4.71 & 20.01   & 2.81    & 12.58   \\
HuatuoGPT *     & 13B         & 24.61  & 12.84  & 7.23   & 4.19   & 7.73 & 27.38   & 7.09    & 17.66   \\
ChatGLM-Med     & \textbf{6B} & 32.18  & 18.37  & 8.87   & 3.79   & 6.09 & 26.14   & 8.08    & 18.87   \\
BenTsao         & 7B          & 32.02  & 17.41  & 8.36   & 3.92   & 6.12 & 17.72   & 3.21    & 14.15   \\
MedChatZH & 7B & \textbf{56.31} & \textbf{32.14} & \textbf{17.58} & \textbf{9.17} & \textbf{10.32} & \textbf{35.99} & \textbf{10.31} & \textbf{21.77} \\ 
\bottomrule\\[-2.5mm]
\multicolumn{8}{l}{$\dagger$ The models highlighted by \textbf{*} means copied scores from HuatuoGPT.}
\end{tabular}%
}
\end{table*}

\section{MedChatZH}
In this section, we will introduce the data process pipeline and training details of MedChatZH. 
\subsection{Data Collection}

Our training dataset consists of two main components: TMC books and raw instructions.

For the medical books, we have gathered a comprehensive collection of over 1,000 books, including renowned works such as the Yellow Emperor's Canon of Internal Medicine and Treatise on Febrile Diseases, as well as valuable folk doctor notes. While we have primarily focused on extracting relevant texts from these books, minimal cleaning has been performed on this dataset.

In contrast, for the instructions component, we have created a mixture of general and medical Chinese data known as med-mix-2M. This dataset combines both general and medical Chinese instructions, providing a diverse range of language patterns and medical contexts. The med-mix-2M dataset serves as a valuable resource for training models with a broad understanding of both general language usage and medical terminology.

\subsection{Data Process Pipeline}

\begin{figure*}[!t]
    \centering
    \includegraphics[width=\textwidth]{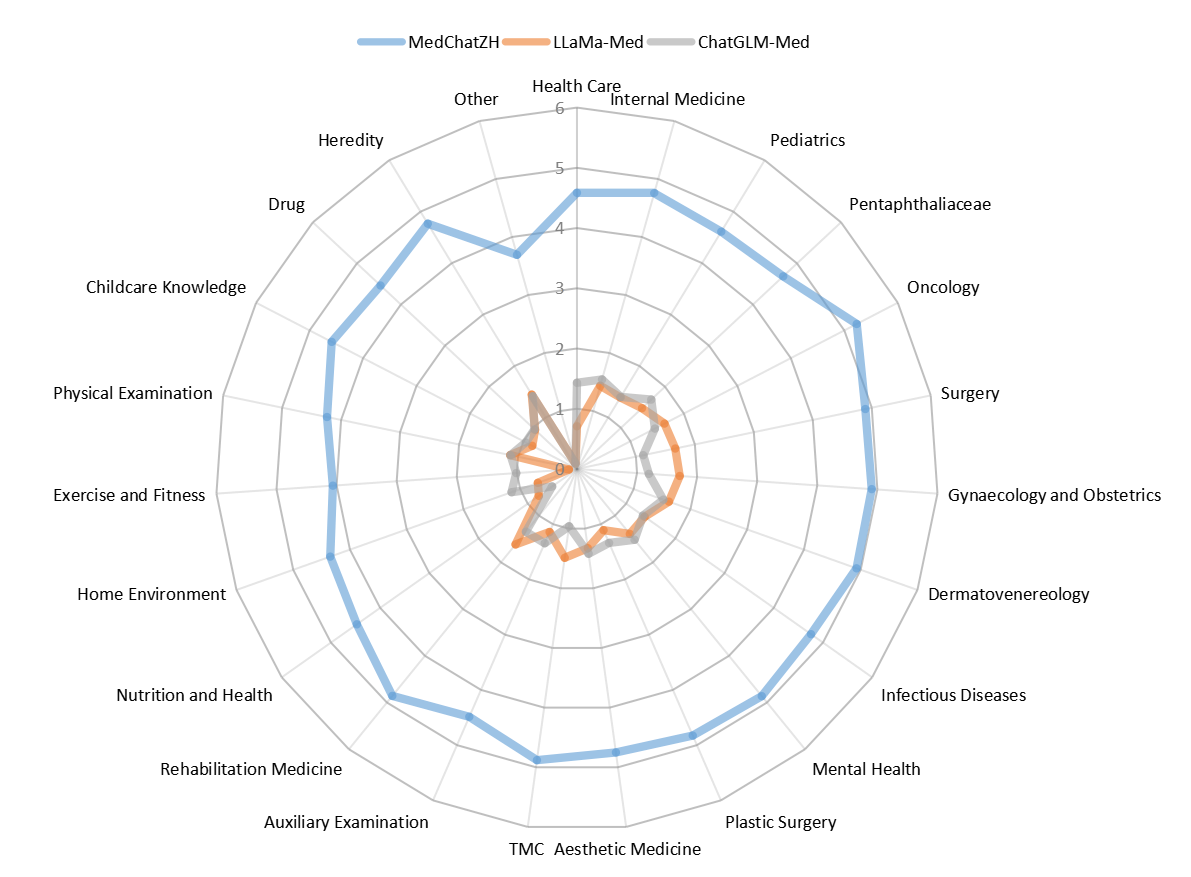}
    \caption{Chinese reward model scores on different categories in Medical QA.}
    \label{fig:Reward model scores}
\end{figure*}

The BELLE-3.5M instruction dataset \cite{belle2023belle} is derived from ChatGPT, employing AI-style instructions known for their high quality. To ensure the dataset's reliability and coherence, we employ heuristic methods during the curation process. Specifically, we discard short answers that consist of fewer than 200 tokens and lack logical consistency. This approach helps to enhance the quality of the question-answer pairs in the dataset, resulting in more accurate and meaningful QA interactions.

To ensure domain-specific knowledge, we have amassed over 7,000,000 medical instructions from the Internet and various Chinese hospitals. These instructions exhibit variations in expression, quality, length, and style. In order to curate a high-quality dataset, we apply the following filtering steps:
\begin{itemize}
    \item Filtering Personal Data: We utilize heuristics, such as regular matching, to identify and remove responses containing personal information like email addresses or phone numbers. This step ensures the protection of individuals' privacy.
    \item Self-labeling and Training: We perform self-labeling on a subset of 3,000 preference ranking data in the medical domain. This subset is then used to train a model called Ziya-LLaMA-7B-Reward \footnote{https://huggingface.co/IDEA-CCNL/Ziya-LLaMA-7B-Reward}. Data with scores lower than 0.5 are discarded, ensuring the selection of high-quality training examples. 
    \item Numerical Symbol Harmonization: We harmonize various numerical symbols, such as '1,', '(1)', etc., into a standardized format represented by a number followed by a dot, e.g., '1.' This standardization ensures consistency and ease of processing for numerical information.
\end{itemize}
As a result of these steps, we obtain a curated dataset comprising 763,629 medical instructions and 1,305,194 general instructions. This dataset serves as the foundation for fine-tuning our model, enabling it to acquire the necessary dialogue capabilities specific to the medical domain.

\subsection{Base Model}

Our base model is Baichuan-7B, which is based on the Transformer and its architecture is the same as the LLaMa. This 7 billion parameter model is trained on about 1.2 trillion tokens supports Chinese and English bilinguals, and the context window length is 4096. The best results of the same size have been achieved on the standard Chinese and English benchmarks (C-Eval/MMLU).

\subsection{Training Details}

Our model is developed using PyTorch 2.0.1, with Baichuan-7B serving as the foundational architecture. During the further pre-training stage, we employ specific settings to optimize the model's performance. The learning rate is set to 2e-5, the batch size per device is 4, and the maximum context length is restricted to 2048 tokens.In the subsequent instruction fine-tuning stage, we deviate from the LoRA \cite{hu2021lora} strategy and instead opt for a full parameter fine-tuning approach. Here, the learning rate is adjusted to 2e-4, the batch size per device is increased to 8, and the maximum context length is limited to 1024 tokens.
For optimization, we employ the AdamW optimizer \cite{loshchilov2017adamw}, and weight decay is set to 1e-5 to mitigate overfitting. To execute our experiments, we utilize 8 NVIDIA A800 GPUs and leverage the ZeRO-2 \cite{rajbhandari2020Zero} stage, which optimizes memory consumption and accelerates training.

\begin{table*}[!t]
\caption{The distribution of the webMedQA dataset is highly skewed, with the largest category being 'internal medicine,' comprising over 17,000 data points. The category with the least representation is 'other,' containing only 30 questions and answers.}
\label{tab:my-table}
\resizebox{\textwidth}{!}{%
\begin{tabular}{@{}lllll@{}}
\toprule
\multicolumn{1}{c}{Dataset Size} & \multicolumn{1}{c}{Count} & \multicolumn{1}{c}{Category}                               &  &  \\ \midrule
\textgreater{}10000 & 2     & Internal Medicine; Surgery             &  &  \\
5000-10000          & 2     & Pediatrics; Gynaecology and Obstetrics &  &  \\
1000-5000 &
  7 &
  \begin{tabular}[c]{@{}l@{}}Pentaphthaliaceae; Oncology; Dermatovenereology; Infectious Diseases; \\ Mental Health; Plastic Surgery; TMC\end{tabular} &
   &
   \\
\textless{}1000 &
  12 &
  \begin{tabular}[c]{@{}l@{}}Health Care; Aesthetic Medicine; Auxiliary Examination; Rehabilitation Medicine; \\ Nutrition and Health; Home Environment; Exercise and Fitness; Physical Examination;\\  Childcare Knowledge; Drug; Heredity; Other\end{tabular} &
   &
   \\ \bottomrule
\end{tabular}%
}
\end{table*}

\section{Experiment}

\subsection{Baselines}

In our evaluation, we compare the performance of our model with that of the state-of-the-art zero-shot model, OpenAI's ChatGPT (GPT-3.5-turbo), as well as several Chinese-specific Language Models (LLMs) that have been fine-tuned specifically on medical domain knowledge.

\begin{itemize}
    \item \textbf{BenTsao} \footnote{https://github.com/SCIR-HI/Huatuo-Llama-Med-Chinese} \cite{wang2023huatuo} is a fine-tuned Chinese Language Model (LLM) developed by SCIR-HI, leveraging the LoRA strategy and Chinese medical knowledge. It consists of two series, LLaMA-7B and Chinese-LLaMA-Alpaca \cite{chinese-llama-alpaca}. Our comparison focuses on LLaMA-7B, which is fine-tuned exclusively on the medical knowledge database, excluding medical literature.
    \item \textbf{ChatGLM-Med} \footnote{https://github.com/SCIR-HI/Med-ChatGLM} is another model based on the same dataset as BenTsao, but utilizing the more Chinese-friendly ChatGLM-6B \cite{du2021glm} as its foundational model. It represents an enhanced version of ChatGLM, specifically designed for improved question-answering effectiveness in the medical field.
    \item \textbf{ChatGPT} \footnote{https://chat.openai.com/} is a sibling model to InstructGPT \cite{ouyang2022instructgpt}, which is trained to follow instructions in a prompt and provide a detailed response. It is considered one of the leading dialogue models, and we compare our model against the GPT-3.5-turbo.
    \item \textbf{HuatuoGPT} \footnote{https://github.com/FreedomIntelligence/HuatuoGPT} \cite{zhang2023huatuogpt} releases model weights of HuatuoGPT-13B, which is trained on Ziya-LLaMA-13B-Pretrain-v1 \footnote{https://huggingface.co/IDEA-CCNL/Ziya-LLaMA-13B-Pretrain-v1}. It combines distilled data from ChatGPT and real-world data from doctors to enhance its medical dialogue capabilities.
\end{itemize}

\subsection{Benchmark}
\begin{figure}[!t]
    \centering
    \includegraphics[width=\linewidth]{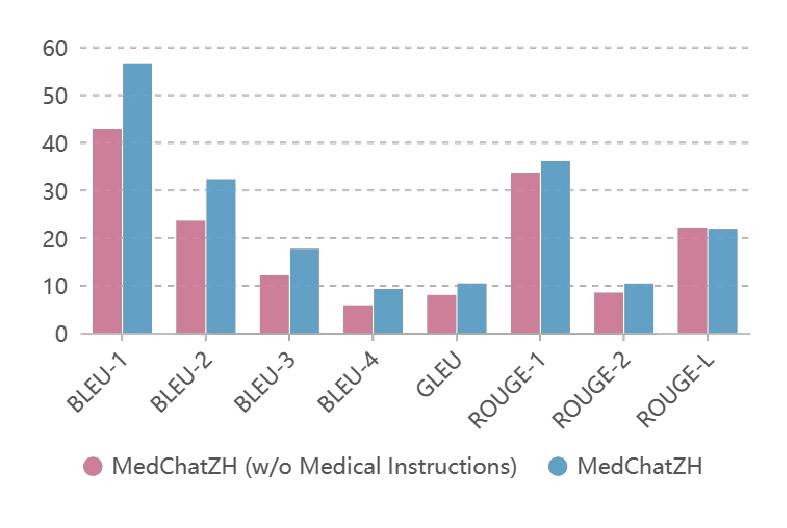}
    \caption{Ablation study on \textbf{webMedQA}, evaluated by traditional NLP metrics.}
    \label{fig:Ablation}
\end{figure}

The \textbf{webMedQA} dataset \cite{he2019webqa} is a real-world collection of Chinese medical question-answering (QA) data sourced from online health consultancy websites. It comprises 63,255 questions\footnote{Note that while HuatuoGPT \cite{zhang2023huatuogpt} states that this dataset contains 63,284 questions, our analysis yielded 63,255 questions.}. This dataset offers the advantage of multiple candidate answers corresponding to each question, allowing for the evaluation of answer accuracy using multiple references. It further categorizes the dataset into 23 different domains, including Health Care, Internal Medicine, and other departments, enabling more targeted analysis and exploration. All basic information can be found in Table 2.

\subsection{Evaluation Metrics}

Our evaluation methodology comprises two primary components: traditional Natural Language Processing (NLP) metrics and reward model scores.

To quantify the similarity between generated and reference sentences, we employ the BLEU metric \cite{papineni2002bleu}. It calculates the k-gram overlap, enabling us to assess the similarity of n-grams in the generated output and the reference sentences.

For evaluating sentence-level fluency, we utilize the GLEU metric \cite{mutton2007gleu}. This metric automatically evaluates the fluency of generated responses, taking into account both adequacy and fluency aspects.

To gauge the overlap of n-grams between the generated output and the reference summaries, we employ the ROUGE metric \cite{lin2004rouge}. Specifically, we employ ROUGE-L, which measures the longest common subsequence of word matches.

Additionally, we incorporate a Reward Model Score as a more flexible and nuanced evaluation metric. In this study, we utilize the Ziya-LLaMA-7B-Reward model. This reward model is specifically designed to accurately assess the quality of model-generated output, including factors such as text repetition, abnormal interruptions, and adherence to instruction requirements. It assigns a lower reward value to outputs that exhibit low-quality generation characteristics.

By combining these traditional NLP metrics and reward-based evaluation, our evaluation framework provides a comprehensive and rigorous assessment of the model's performance. These metrics enable us to evaluate similarity, fluency, adherence to instructions, and overall quality of the generated responses in a systematic and objective manner.

\subsection{Results}

In this research study, our primary focus is on evaluating single-turn questions. The results of all the models are presented in Tab 1. It's important to note that the score results for GPT-3.5-turbo and HuatuoGPT are directly taken from the original paper of HuatuoGPT, and we have not re-run the experimental validation for these models. However, for the remaining models, we have used official checkpoints and conducted inferences on the dataset to ensure that all results are reproducible. Our model demonstrates a significant performance improvement over other baseline models in Single-turn Chinese medical dialogue situations.

Due to the limitations of traditional metrics commonly used in machine translation scenarios, which may not be entirely suitable for evaluating dialogue quality, we have also employed a fine-tuned reward model to score answers. For this purpose, we utilized a medical-specific language model in the Chinese domain to compare the performance of our model against other baselines, as shown in Fig 1.

To ensure accurate evaluation and avoid unnecessary confusion, it is essential to consider that different versions of the evaluation kit can yield different results \cite{shi2022evaluation}. Therefore, we have used the latest version of NLTK-3.8.1 for our evaluation.

\section{Discussion}

\subsection{Ablation Study}

Given the constraints imposed by limited computational resources, we have conducted an ablation study focusing solely on whether to use distilled medical instructions. The results, as depicted in Fig 2, clearly demonstrate that after fine-tuning the model using high-quality medical instruction data, the medical question-answering (QA) ability has shown a substantial improvement. This outcome highlights the crucial role played by fine-tuning with relevant medical instruction data in enhancing the performance of our model in the medical QA domain.

\subsection{Limitation}

Our model is trained for Chinese speakers in the non-commercial medical domain, so it's not suitable for other languages or domains. Medical advice is sensitive and critical, and if the model provides unreasonable advice, it could lead to bad negative effects. We cannot guarantee the authenticity of our model's output, and it may suffer from hallucination phenomena common in language models. Caution, human verification, and transparent communication are essential when using the model.

\section{Conclusion}

This paper compiles and organizes a significant amount of traditional Chinese medicine texts to further train Chinese large models. This process enhances the models' localization and adaptability to specific language environments. Additionally, the data quality is improved through a rigorous Data cleansing process that involves heuristic methods and reward models.

To evaluate the effectiveness of the approach, comprehensive tests were performed using real medical consultation data. These tests compared MedChatZH with multiple powerful baselines, including traditional NLP indicators and AI model scoring. The results demonstrate the robustness of MedChatZH in the medical domain, validating its performance and efficacy.

\section{Acknowledgements}

Supported by Research Programme of National Engineering Laboratory for Big Data Distribution and Exchange Technologies,  Shanghai Municipal Special Fund for Promoting High Quality Development (No. 2021-GYHLW-01007)

% Entries for the entire Anthology, followed by custom entries
\bibliography{anthology,custom}
\bibliographystyle{acl_natbib}

\appendix

\end{document}